\documentclass{article}

\usepackage{microtype}
\usepackage{multirow}
\usepackage{array}
\usepackage{subfigure}
\usepackage{booktabs} 
\usepackage[utf8]{inputenc} 
\usepackage[T1]{fontenc}    
\usepackage{hyperref}       
\usepackage{url}            
\usepackage{booktabs}       
\usepackage{amsfonts}       
\usepackage{nicefrac}       
\usepackage{microtype}      

\usepackage{comment}
\usepackage{tikz}
\usepackage{comment}
\usepackage{amsmath,amssymb} 
\usepackage{color}
\usepackage{wrapfig}

\usepackage{amsmath}
\usepackage{amssymb}
\usepackage{booktabs}
\usepackage{url}
\usepackage{makecell}
\usepackage{floatrow}
\usepackage{epsfig}
\usepackage{graphicx}
\usepackage[titletoc,title]{appendix}
\usepackage{comment}
\usepackage{array}
\usepackage{multirow}
\usepackage{enumitem}
\usepackage{wrapfig}
\usepackage{graphicx}
\usepackage{lipsum}
\usepackage{tabularx}
\usepackage{wrapfig,lipsum,booktabs}
\usepackage{algorithm}
\usepackage{algorithmic}

\usepackage[accsupp]{axessibility}
\usepackage{hyperref}

\usepackage[accepted]{icml2023}

\usepackage{amsmath}
\usepackage{amssymb}
\usepackage{mathtools}
\usepackage{amsthm}
\usepackage{bbm}
\usepackage{soul}

\usepackage[capitalize,noabbrev]{cleveref}

\theoremstyle{plain}

\theoremstyle{definition}

\theoremstyle{remark}

\usepackage[textsize=tiny]{todonotes}

\newcommand{\model}{\text{Slot-TTA}}

\newcommand{\CNN}{\mathrm{CNN}}
\newcommand{\Enc}{\mathrm{Enc}}
\newcommand{\Dec}{\mathrm{Dec}}

\newcommand{\I}{\mathrm{I}}

\newcommand{\maskformerbsln}{Mask2Former}
\newcommand{\maskd}{Mask3D}
\newcommand{\maskformerbyolbsln}{Mask2Former-BYOL}
\newcommand{\maskformerreconbsln}{Mask2Former-Recon}

\icmltitlerunning{Test-time Adaptation with Slot-Centric Models}

\begin{document}

\twocolumn[
\icmltitle{Test-time Adaptation with Slot-Centric Models}



\icmlsetsymbol{equal}{*}

\begin{icmlauthorlist}
\icmlauthor{Mihir Prabhudesai}{yyy}
\icmlauthor{Anirudh Goyal}{comp}
\icmlauthor{Sujoy Paul}{sch}
\icmlauthor{Sjoerd van Steenkiste}{sch}
\icmlauthor{Mehdi S. M. Sajjadi}{sch}
\icmlauthor{Gaurav Aggarwal}{sch}
\icmlauthor{Thomas Kipf}{sch}
\icmlauthor{Deepak Pathak}{yyy}
\icmlauthor{Katerina Fragkiadaki}{yyy}
\end{icmlauthorlist}

\icmlaffiliation{yyy}{Carnegie Mellon University}
\icmlaffiliation{comp}{Mila, DeepMind}
\icmlaffiliation{sch}{Google Research}
\icmlcorrespondingauthor{Mihir Prabhudesai}{\href{mailto:mprabhud@cs.cmu.edu}{mprabhud@cs.cmu.edu}}

\icmlkeywords{test-time adaptation, object-centric learning}

\vskip 0.3in
]



\printAffiliationsAndNotice{}  


\begin{abstract}
Current visual detectors, though impressive within their training distribution, often fail to parse out-of-distribution scenes into their constituent entities.
Recent test-time adaptation methods use auxiliary self-supervised losses to adapt the network parameters to each test example independently and have shown promising results towards generalization outside the training distribution for the task of image classification. In our work, we find evidence that these losses are insufficient for the task of scene decomposition, without also considering architectural inductive biases. 
Recent slot-centric generative models attempt to decompose scenes into entities in a self-supervised manner by reconstructing pixels. Drawing upon these two lines of work, we propose \model{}, a semi-supervised slot-centric scene decomposition model that at test time is adapted \textit{per scene} through gradient descent on reconstruction or cross-view synthesis objectives. 
We evaluate \model{} across multiple input modalities, images or 3D point clouds, and show substantial out-of-distribution performance improvements against state-of-the-art supervised feed-forward detectors,  and alternative test-time adaptation methods. 
Project Webpage: \url{http://slot-tta.github.io/}
\end{abstract}

\section{Introduction}

\label{sec:intro}
While significant progress has been made in scene perception within the last decade, decomposing scenes into familiar entities often generalizes poorly outside the training distribution \citep{geirhos2020shortcut,hendrycks2021many}. 

To tackle changes in the data distribution, Test-Time Adaptation (TTA) methods~\citep{ghifary2016deep,sun2020test,wang2020tent}  adapt the model parameters at test-time to help generalization. In recent years, a variety of methods based on TTA have been proposed, focusing on few-shot adaptation~\citep{ren2018meta} where the network is given access to a few \emph{labeled} examples, or unsupervised domain adaptation (UDA)~\citep{zhang2021survey} where the network is given access to many \emph{unlabelled} examples from the new distribution. A popular approach in this setting is pseudo-labelling~\citep{wang2020tent,bateson2022test}, where the network uses its confident predictions in some examples as additional pseudo-labelled training data to improve its accuracy. However, this approach requires \textit{multiple} confident examples for adaptation.

\begin{figure}[t!]
  \centering
    \includegraphics[width=1.0\textwidth]{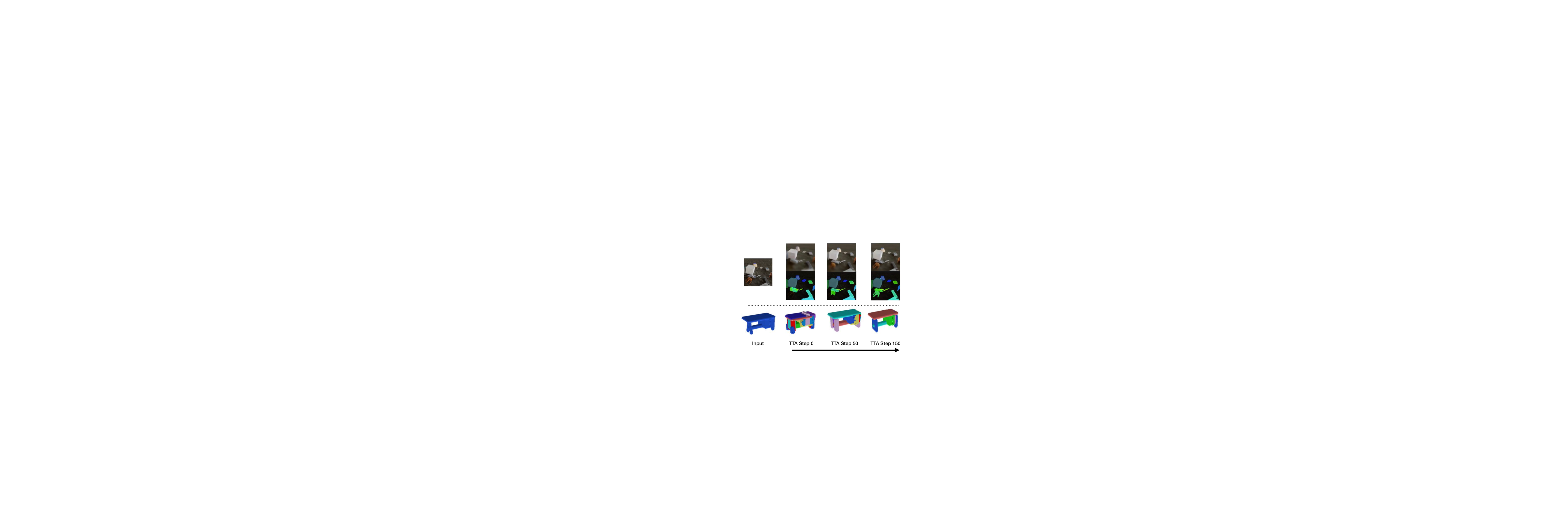}
    \caption{ \textbf{Test-time adaptation in \model{}}: Segmentation improves when optimizing reconstruction or view synthesis objectives via gradient descent at test-time on a single test sample.} 
  \label{fig:slowinfer}
\end{figure}

We instead, study a specific unsupervised domain adaptation (UDA ) setting  where the network is adapted \textit{independently} to each unlabelled example in the test set. This setting is analogous to a human taking more time to parse a difficult example while \emph{not} having access to any additional information \citep{kahneman2011thinking}.

Existing approaches in this setting typically devise a loss for a self-supervised pre-text task, such as rotation prediction 
 in TTT~\citep{sun2020test} or instance discrimination in MT3~\citep{bartler2022mt3},  and then optimize this loss per image at test-time ~\citep{sun2020test,maettt,bartler2022mt3,grill2020bootstrap}. While these methods have demonstrated success for the task of image classification, 
 they are not equally effective when applied to other tasks, such as scene decomposition, that requires reasoning about objects, as we show in experiment Section \ref{sec:multiview}. Specifically, when we apply the instance discrimination loss of MT3~\citep{bartler2022mt3} to the state-of-the-art scene segmentation model \maskformerbsln{} of \citet{mask2former}, the segmentation performance deteriorates during test-time adaptation. The question then becomes: what would be an effective TTA method for the task of scene decomposition? 
 
Recent slot-centric generative models that attempt to segment scenes into object entities completely unsupervised, by optimizing a reconstruction objective~\citep{eslami2016attend,greff2016tagger,van2018relational,goyal2019recurrent,kosiorek2019stacked,locatello2020object,zoran2021parts} share the  end-goal of scene decomposition and thus become a good candidate architecture for TTA. These methods differ in details but share the notion of incorporating a fixed set of entities, also known as \emph{slots} or \emph{object files}. Each slot extracts information about a single entity during encoding and is ``synthesized'' back to the input domain during decoding. 

In light of the above, we propose Test-Time Adaptation with Slot-Centric models ($\model$), a semi-supervised model equipped with a slot-centric bottleneck \citep{locatello2020object} that jointly \emph{segments}  and \emph{reconstructs} scenes. 
At training time, $\model$ is trained supervised to jointly \emph{segment}  and \emph{reconstruct} 2D (multi-view or single-view) RGB images or 3D point clouds.
At test time, the model adapts to a single test sample by updating its network parameters solely by optimizing the reconstruction objective through gradient descent, as shown in Figure \ref{fig:slowinfer}. 
\model{} builds on top of slot-centric models by incorporating segmentation supervision during the training phase. 
Until now, slot-centric models have been neither designed nor utilized with the foresight of Test-Time Adaptation (TTA). In particular, \citet{engelcke2020reconstruction} showed that TTA via reconstruction in slot-centric models fails due to a reconstruction-segmentation trade-off: as the entity bottleneck loosens, there's an improvement in reconstruction; however, segmentation subsequently deteriorates. 
We show that segmentation supervision aids in mitigating this trade-off and helps scale to scenes with complicated textures.
 We show TTA in semi-supervised slot-centric models significantly improves scene decomposition.

We test \model{} in scene segmentation of multi-view posed images, single-view images and 3D point clouds  in the datasets of PartNet \citep{mo2019partnet}, MultiShapeNet-Hard \citep{sajjadi2022scene} and CLEVR \citep{johnson2017clevr}. The model segments objects and parts while reconstructing them in 2D or 3D. We compare its segmentation performance against state-of-the-art supervised feedforward RGB image and 3D point cloud segmentors of \maskformerbsln{} and \maskd{}~\cite{mask2former,schult2022mask3d}, NeRF-based multi-view segmentation fusion methods of \cite{zhi2021place} that  adapt per scene through RGB and segmentation rendering,  state-of-the-art test-time adaptation methods \citep{bartler2022mt3}, unsupervised entity-centric generative models \citep{locatello2020object,sajjadi2022object}, and semi-supervised 3D part detectors \citep{Wu_2020_CVPR,shap2prog}. 
We show that \model{} outperforms SOTA feedforward  segmentators in out-of-distribution scenes, dramatically outperforms alternative TTA methods and alternative unsupervised or semi-supervised scene decomposition methods \citep{locatello2020object,sajjadi2022object,Wu_2020_CVPR,shap2prog}, and better exploits multi-view information for improving segmentation over semantic NeRF-based multi-view fusion.
Additionally, we show that test-time adaptation not only improves segmentation accuracy but also enhances the rendering quality of novel (unseen) views that were \textit{not} used during test-time training.

Our contributions are as follows:

(i) We present an algorithm that significantly improves scene decomposition accuracy for out-of-distribution examples by performing test-time adaptation on each example in the test set independently.

(ii) We showcase the effectiveness of SSL-based TTA approaches for scene decomposition, while previous self-supervised test-time adaptation methods have primarily demonstrated results in classification tasks.

(iii) We introduce semi-supervised learning for slot-centric generative models, and show it can enable these methods to continue learning during test time. In contrast, previous works on slot-centric generative have neither been trained with supervision nor been used for test time adaptation.

(iv) Lastly, we devise numerous baselines and ablations, and evaluate them across multiple benchmarks and distribution shifts to offer valuable insights into test-time adaptation and object-centric learning.

Our code is publicly available to the community on our project webpage: \url{http://slot-tta.github.io}.

\section{Related  Work}

\paragraph{Test-time adaptation}
In test-time adaptation, model parameters are updated at test-time for the model to better generalize to data distribution shifts. In recent years, there has been significant development in this direction. Methods such as pseudo labelling and entropy minimization~\citep{shin2022mm,wang2020tent, iwasawa2021test, bateson2022test} have demonstrated that supervising the model using its confident predictions  helps improve its accuracy. Adaptive BatchNorm methods~\citep{khurana2021sita,chang2019domain} have shown that updating the BatchNorm parameters using a set of examples can help adaptation. 
Despite their impressive performance, these methods inherently require confident predictions or a batch of examples to adapt. Self-supervised learning (SSL)~\citep{sun2020test,bartler2022mt3,maettt} based TTA methods on the other hand, train using a combination of the task and a SSL loss. During test time, they optimize using only the SSL loss. They can adapt to \textit{individual} examples at test time.  However, all methods in the SSL setting thus far focus on the image classification task and mainly differ in terms of the SSL loss employed. For example TTT~\citep{sun2020test} uses rotation angle prediction as their SSL loss, MT3~\citep{bartler2022mt3} uses a BYOL~\citep{grill2020bootstrap} loss and  TTT-MAE~\citep{maettt} uses Masked autoencoding loss ~\citep{pathak2016context,he2022masked}. Our work targets TTA for the task of scene decomposition. In our work, we show that TTA with reconstruction loss in slot-centric models can help improve the segmentation peformance in out-of-distribution scenes. 

\paragraph{Slot-centric generative models for scene decomposition}
\textit{Entity-centric} (or \textit{object-centric}) models represent a visual scene in terms of separate object variables, often referred to as \textit{slots} or \textit{object files}~\citep{greff2020binding, sabour2017dynamic,kosiorek2018sequential,engelcke2019genesis, goyal2020object, ke2021systematic, burgess2019monet,greff2019multi,zablotskaia2020unsupervised,rahaman2020s2rms}. Prominent examples of such models include MONet~\citep{burgess2019monet}, GENESIS \citep{engelcke2019genesis}, IODINE \citep{greff2019multi}, and Slot Attention (SA) \citep{locatello2020object}, which are trained in a fully-unsupervised setting via a simple auto-encoding objective. Scene decomposition emerges via the inductive bias of the model architecture (and in some cases, additional regularizers). 
OSRT \citep{sajjadi2022object} builds on top of SA by replacing their autoencoding objective with a novel-view synthesis objective. \model{} builds on top of slot-centric models by adding segmentation supervision at training time, and using reconstruction optimization per example for TTA.

\section{Method}

\begin{figure*}[t!]
  \centering
    \includegraphics[width=0.95\textwidth]{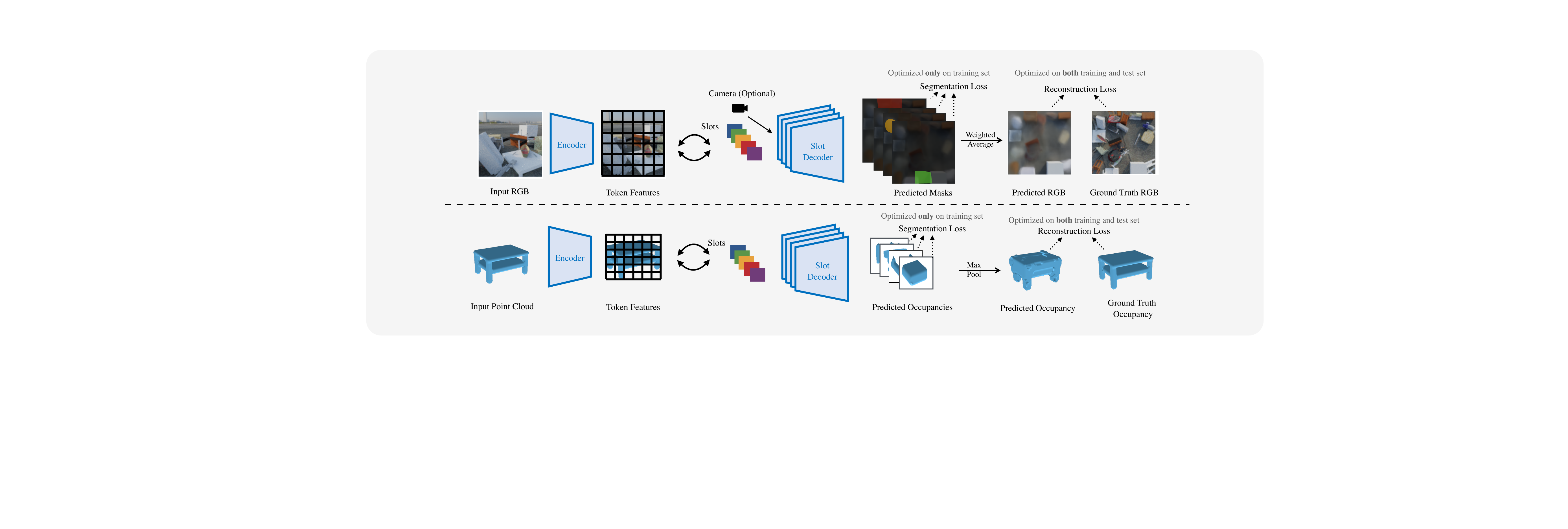}
    \vspace{-0.5cm}
    \caption{\textbf{Model architecture for \model{} for posed multi-view or single view RGB images (\textit{top}) and 3D point clouds (\textit{bottom})}. 
    \model{} maps the input (multi-view posed) RGB images or 3D point cloud to a set of token features with appropriate encoder backbones. It then maps these token features  to a set of slot vectors using Slot Attention. Finally, it  decodes each slot into its respective segmentation mask and RGB image or 3D point cloud. It  uses weighted averaging or maxpooling to fuse renders across all slots. For RGB images, we show results for multi-view and single-view settings, where in the multi-view setting the decoder is conditioned on a target camera-viewpoint. We train \model{} using reconstruction and segmentation losses. At test time, we optimize only the reconstruction loss. }
  \label{fig:semisupervised}
\end{figure*}

The goal of \model{} is to decompose scenes into objects or parts. 
We consider three different settings: (i) 2D multi-view RGB images (ii) 2D single-view RGB images and (iii) 3D point clouds. 
In each setting, the model encodes the scene as a set of slot vectors that capture information about individual objects and decodes them back to either (novel-view) RGB images or 3D point clouds, depending on the setting. 
To compute slots,  \model{} uses Slot Attention (SA) \citep{locatello2020object}, where visual features are softly partitioned across slots through iterative attention.

\subsection{Background} \label{sec:background}
Current state-of-the-art detectors and segmentors instantiate slots, a.k.a. query vectors, from 2D visual feature maps or 3D point feature clouds \citep{mask2former,schult2022mask3d} via iterative cross-attention (features to slots) and self-attention (slots to slots) operations \citep{carion2020end,mask2former}. 
Recently, Slot Attention of \citet{locatello2020object} and Recurrent Independent Mechanisms (RIMs) of \citet{goyal2019recurrent} popularized  competition amongst slots and iterative routing to encourage a single location in the input to be assigned to a unique slot vector.

Given a set of feature vectors $M \in \mathbb{R}^{N \times C}$ obtained from an encoder, where $N$ is the number of tokens in the encoded feature map and $C$ is the dimensionality of each token. The Slot Attention module compresses these $N$ tokens into a set of $P$ slot vectors  $S \in \mathbb{R}^{P \times D}$, where $D$ is the dimensionality of each slot vector.

It does so by updating a set of learned latent embedding vectors $\hat{S} \in \mathbb{R}^{P \times D}$, conditioned on $M$. Specifically it computes an attention matrix $A \in \mathbb{R}^{N \times P}$ between  feature map $M$ and $\hat{S}$ using the equation \ref{eq:softmax}
\begin{equation}
    A_{i,p} = \frac{\exp{(k(M_i) \cdot q(\hat{S_p})^T)}}{\sum_{p=0}^{P-1}\exp{(k(M_i) \cdot q(\hat{S_p})^T)}} 
    \label{eq:softmax}
\end{equation}

$$
\hat{M} = v(M)
$$

here  $k$, $q$, and $v$ are learnable linear transformations. Specifically $k$ and $v$ are applied element-wise to map $M \in \mathbb{R}^{N \times C}$ to $\mathbb{R}^{N \times D}$. Similarly, $q$ applies an element-wise transformation from $\hat{S} \in \mathbb{R}^{P \times D}$ to $\mathbb{R}^{P \times D}$. 
The softmax normalization over the slot axis in equation 1 ensures competition amongst them to attend to a specific feature vector in $M$. It then extracts slot vectors $S$ from $M$ by updating $\hat{S}$ using a GRU. 

$S = \mathrm{GRU}(\hat{S}, U)$ where the update vector $U$ is calculated by taking a weighted average of elements in $\hat{M}$ using the re-normalized attention matrix $\hat{A}$:

 $$ U = \hat{A}^T \hat{M} 
 \in \mathbb{R}^{P \times C}, \textrm{where}  \  \hat{A}_{i,p} = \dfrac{{A}_{i,p}}{\sum_{i=0}^{N-1}A}_{i,p} $$

 We iterate 3 times over equations 1 to 4, while setting $\hat{S}$ = $S$ each time.

\subsection{Test-time Adaptation with Slot-Centric Models (\model)} \label{sec:gfs}

We first describe the encoders and decoders of \model{} for each modality. Next, we detail how we train \model{} and how we  adapt it at test time. 

\subsubsection{Encoding and Decoding Backbones}
\label{sec:encdec}
\paragraph{Posed multi-view 2D RGB images}
The architecture of \model{} for the the multi-view RGB setting is illustrated in Figure \ref{fig:semisupervised} \textit{top}. 
Our model's architecture is built upon OSRT   \citep{sajjadi2022object}, which is an object-centric, geometry-free novel view synthesis method.    
Given a set of  posed RGB images as input, a CNN encodes each input image $\I_i$ into a feature grid, which is then flattened into a set of tokens with camera pose and ray direction information added in each of the tokens, similar to SRT \citep{sajjadi2022scene}. These are then encoded into a set of latent features using a transformer~\citep{vaswani2017attention} $\Enc$ with multiple self-attention blocks $M = \Enc(\CNN(\I_i))$.  The latent features $M$ are then mapped into a set of slots $S$ using Slot Attention (Section \ref{sec:background}). 

For decoding, we use a spatial broadcast decoder~\citep{watters2019spatial}, where a render MLP takes as input the slot vector $S_k$ and the pixel location $p$ parameterized by the camera position and the ray direction of the pixel to be decoded, and  outputs an RGB color $c_k$ and an unnormalized alpha score $a_k$ for each pixel location $c_k,a_k = \Dec(p,S_k)$. The $a_k$'s are normalized using a Softmax and used as weights to aggregate the predicted RGB values $c_k$ for each slot. 
We ablate other decoder choices, such as the Slot Mixer decoder \citep{sajjadi2022object} in Appendix Section \ref{sec:results_rgbmulti_sup}.

\paragraph{Single-view 2D RGB images}

The pipeline of \model{} for the the single-view RGB setting is the same as the multi-view setting (Figure \ref{fig:semisupervised} top), except we do not condition the decoder with the camera information. 
In this setting, $\model$ uses a convolutional encoder the same as \citet{locatello2020object}, to encode the input RGB image into a feature grid. We then add positional vectors to the feature grid and map them to a set of slot vectors using Slot Attention. Similar to the multi-view setting, each slot vector is decoded to the RGB image and an alpha mask using an MLP renderer. We parameterize pixel location $p$ as $(x,y)$ points on the grid instead of camera information.

\paragraph{3D point clouds}
The architecture of \model{} for the 3D point cloud setting is illustrated in Figure \ref{fig:semisupervised} \textit{bottom}.  
Our model's  architecture uses a 3D point transformer  \citep{zhao2021point} which maps the 3D input points to a set of $M$ feature vectors of $C$ dimensions each. We set $M$ to 128 and $C$ to 64 in our experiments. 
Point feature vectors are mapped to slots with Slot Attention. 
$\model$ decodes 3D point clouds from each slot  using  implicit functions \citep{mescheder2019occupancy}. Specifically, each decoder takes in as input the slot vector $S_k$ and an $(X,Y,Z)$ location and returns the corresponding occupancy score $o_{k,x,y,z} = \Dec(S_k, (x,y,z))$, where $\Dec$ is a multi-block ResNet MLP similar to that of \citet{coconets}. 
We then max-pool over the slot dimension $k$ to get an occupancy value $o_{x,y,z}$ for each 3D point in the scene.
More details on our encoder and decoder architectures are included in the Appendix  Section \ref{sec:impl}.

\paragraph{Information bottleneck in the decoder}
A very important ingredient for scene decomposition via optimizing reconstruction  in slot-centric models is the information bottleneck in the decoder \citep{engelcke2020reconstruction,locatello2020object,sajjadi2022object}. 
In slot-centric models, the decoder  $\Dec(S_k, (x,y,z))$ decodes the segmentation mask \textit{conditioned only on $S_k$},  where $S_k$ is a $C$ dimensional slot vector, whereas  in supervised image segmentors \citep{mask2former,carion2020end}  the decoder  $\Dec(S_k, M)$ decodes the segmentation mask \textit{conditioned on both the slot vector $S_k$ and the feature map $M$.}  Specifically, in the latter case, they use a dot product between the vectors in the encoded feature map  $M$ and slot vectors $S_k$, thus not having an information bottleneck. These different decoder choices create interesting trade-offs between test-time adaptation and fitting well to the training distribution.  We discuss this  further  in Section \ref{sec:lim}. 
\vspace{-0.1in}
\subsubsection{Training and Test-time Adaptation}
 \model{} is supervised from entity segmentation masks and self-supervised from image and 3D point cloud reconstruction  objectives.

 \paragraph{Training for joint segmentation and reconstruction}
\label{sec:joint}
Our model is trained to jointly optimize a (novel view) image synthesis or point cloud reconstruction objective alongside  a task-specific segmentation objective over all the $n$ examples in the training set. The optimization reads: 
\begin{equation}
    \min_{\theta} \frac{1}{n}\sum_{i=1}^{n} \lambda_{s} l_{seg}(x_i, y_i; \theta) + \lambda_{r} l_{recon}(x_i; \theta), \label{eq:loss}
\end{equation}
where $x$ represents the input scene and $y$ the segmentation labels. 
For RGB image reconstruction, we minimize the mean squared error between the predicted and ground truth RGB images. For segmentation, we supervise the alpha masks $a_i$ of each slot as provided by the decoders. 
We use Hungarian matching~\citep{kuhn1955hungarian}, a combinatorial optimization algorithm that solves assignment problems, to associate the ground truth masks with the predicted masks, and upon association we apply a categorical cross-entropy loss $l_{seg}$.  
For 3D point cloud reconstruction, we supervise the predicted occupancy probability $o$. We use a binary cross-entropy loss for $l_{recon}$. For $l_{seg}$ we use Hungarian matching with a categorical cross-entropy loss over $o_k$. We weight the respective losses by $\lambda_s$ and $\lambda_r$.

\paragraph{Test-time adaptation} We refer to a single forward pass through our trained model without any test-time adaptation as \textit{direct inference} (same as regular inference). 
During test-time adaptation, we adapt the parameters $\theta$ of our model by backpropagating through \textit{only} the reconstruction objective of Eq.\ref{eq:loss} for 150 steps per scene example. For the multi-view posed image case, we test-time adapt the model considering RGB reconstruction on target RGB views, and measure segmentation performance similarly on the same target views (that are different from the input views). 
We visualize the  test-time adaptation results across variable number of iterations in Figure \ref{fig:slowinfer}.
Further in our appendix Section \ref{sec:results_rgbmulti_sup} we ablate different choices of parameters to update during TTA.

\section{Experiments}\label{sec:experiments}
We test $\model$ in its ability to segment multi-view posed RGB images, single-view RGB images and 3D point clouds. Further, we test $\model$'s ability to render and decompose image views from  novel (unseen) viewpoints.
Our experiments aim to answer the following questions: 
\begin{itemize}
\item How does $\model{}$ compare against state-of-the-art 2D and 3D segmentors, Mask2Former \citep{mask2former} and Mask3D  
 \cite{schult2022mask3d}, within and outside of the training distribution?
\item How does $\model{}$ compare against previous state-of-the-art test-time adaptation methods \citep{bartler2022mt3}?
\item How does $\model{}$ compare against NeRF-based methods  that do multi-view semantic fusion \citep{zhi2021place}? 
\item How much do different design choices of our model contribute to performance? We investigate decoder architecture, mask segmentation supervision, and the use of Slot Attention.
\end{itemize}

We use Adjusted Random Index (ARI) as our segmentation evaluation metric ~\citep{rand1971objective}. ARI measures cluster  similarity while being invariant to the ordering of the cluster centers. ARI of 0 indicates random clustering, while 1 indicates a perfect match. Note that we \textit{do include the background component} in our ARI metric. We use the publicly available implementation of~\citet{multiobjectdatasets19}.

\subsection{Decomposing RGB images in multi-view scenes}
\vspace{-0.1cm}
\label{sec:multiview}
\begin{figure}[h!]
  \centering
    \includegraphics[width=1.0\textwidth]{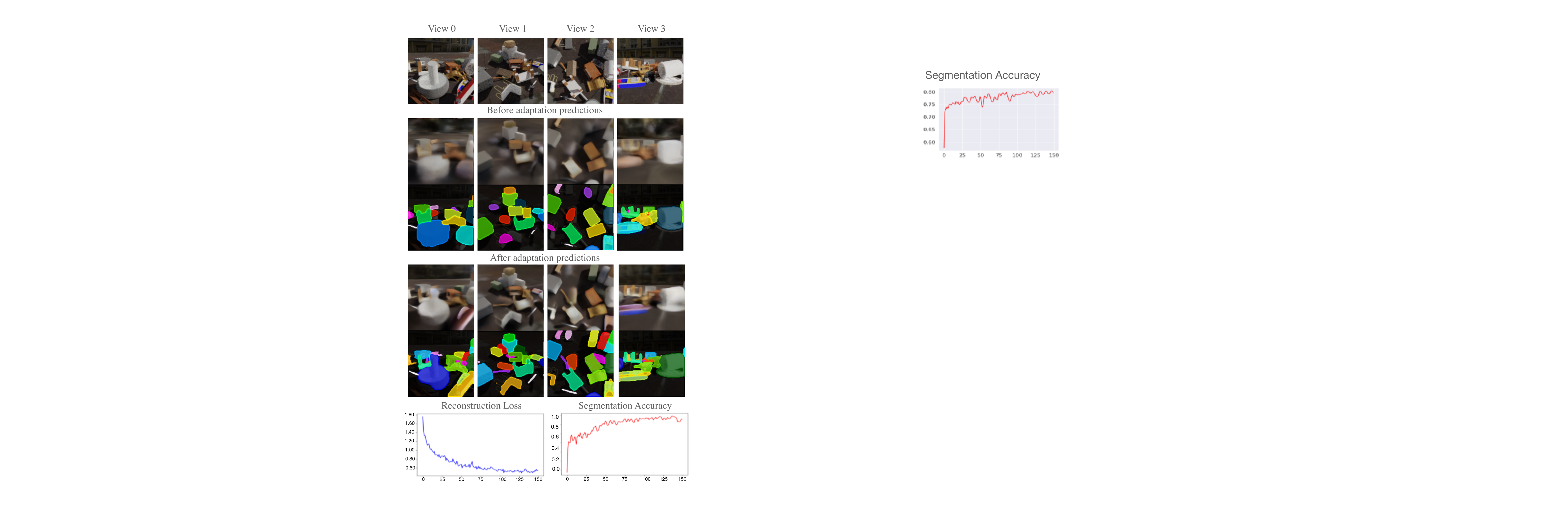}
    \vspace{-0.8cm}
    \caption{\textbf{Test-time adaptation in \model{} for multi-view RGB images}.  
    We visualize image reconstruction loss (blue curve) and  segmentation ARI accuracy (red curve) during TTA iterations. 
     Segmentation accuracy increases as reconstruction loss decreases. 
    }
  \label{fig:multiview_tta}
\end{figure}

\paragraph{Dataset}
We evaluate \model{} on the MultiShapeNet-Hard (MSN) dataset from SRT of \cite{sajjadi2022scene}. The dataset is constructed by rendering 51K ShapeNet objects using Kubric~\citep{greff2022kubric} simulator against 382 real world HDR backgrounds. Each scene has 9 posed RGB rendered images that are randomly assigned into input and target views for our model.
We consider a  train-test split where 
we ensure that \textit{there is no overlap between object instances and between number of objects present in the scene} between training and test sets.
Specifically, scenes with 5-7 object instances are in the training set, and scenes with 16-30 objects are in the test set. 
Increasing the amount of clutter or occlusions in the scene has shown to be a common test for a model's strong generalization~\citep{cai2020messytable}.  
In the Appendix,  
we test our model on a different distribution shift where  we introduce instances from \textit{unseen object categories} from Google Scanned objects dataset \cite{downs2022google} in the test set (Table \ref{tab:gso}). Further, in appedix Table \ref{tab:clevr}, we evaluate \model{} on multi-view CLEVR dataset \cite{johnson2017clevr}.

\paragraph{Baselines}
We consider the following  baselines: 

(i) \maskformerbsln{} \citep{mask2former}, a state-of-the-art 2D image segmentor  that extends  detection transformers \citep{carion2020end} to the task of image segmentation via using multiscale segmentation decoders with masked attention. 

(ii) \maskformerbyolbsln{} which combines the segmentation model of \citet{mask2former} with test time adaptation using BYOL self-supervised loss of \citet{bartler2022mt3}. 

(iii) \maskformerreconbsln{} which combines the segmentation model of \citet{mask2former} with a RGB rendering module and an image reconstruction objective for test-time adaptation. 

(iv) Semantic-NeRF \citep{zhi2021place}, a NeRF model which adds a segmentation rendering head to the multi-view RGB rendering head of traditional NeRFs. It is fit per scene \textit{on all available 9} RGB posed images and corresponding segmentation maps from \maskformerbsln{} as input. 

(v) \model{}-w/o supervision, a variant of our model that does not use any segmentation supervision; rather is trained only for cross-view image synthesis  similar to OSRT \citep{sajjadi2022object}.

\begin{table}[t]
\centering
\scalebox{0.68}{
\begin{tabular}{@{\extracolsep{0pt}}llcccc@{}}
\toprule
\multirow{2}{*}{\textbf{Method}}
&\multicolumn{2}{c}{\textbf{in-dist : 5-7 instances}} 
&\multicolumn{2}{c}{\textbf{out-dist : 16-30 instances}} \\
\cmidrule{2-3} \cmidrule{4-5}
 & Direct Infer.  & with TTA.  & Direct Infer.  & with TTA.  \\
\midrule
\model{}-w/o supervision    & \quad 0.32 & 0.30 & 0.33 &  0.29 \\
\midrule  
\maskformerbsln & \quad 0.93 & N/A & 0.74 & N/A \\
\maskformerbyolbsln & \quad 0.93 & \textbf{0.95} & 0.75 & 0.74 \\
\maskformerreconbsln & \quad 0.93 & 0.92 & 0.74 & 0.67 \\
\midrule
Semantic-NeRF & \quad N/A & 0.94 & N/A & 0.77 \\
\midrule
\model{} (Ours)   & \quad 0.92 & \textbf{0.95} & 0.70 &  \textbf{0.84} \\
\bottomrule
\end{tabular}
}
\caption{\textbf{Instance Segmentation ARI accuracy} (higher is better) in the multi-view RGB setup for in-distribution test set of 5-7 object instances and out-of-distribution 16-30 object instances.
}
\label{tab:osrt}
\end{table}

\paragraph{Results}
We show quantitative segmentation results for our model and baselines on target camera viewpoints in Table \ref{tab:osrt} and qualitative TTA results in Figure \ref{fig:multiview_tta}.  
Our conclusions are as follows:

(i) \model{} with TTA outperforms \maskformerbsln{} in  out-of-distribution scenes and has comparable performance within the training distribution. 

(ii)  \maskformerbyolbsln{} does not improve over \maskformerbsln{}, which suggests that adding self-supervised losses of SOTA image classification TTA  methods \citep{bartler2022mt3} to scene segmentation methods does not help.

(iii) \model{}-w/o supervision (model identical to 
 \citet{sajjadi2022object}) greatly underperforms a supervised segmentor \maskformerbsln{}. This means that unsupervised slot-centric models are still far from reaching their supervised counterparts. 

(iv)  \model{}-w/o supervision does not improve during test-time adaptation.  This suggests \textit{segmentation supervision at training time is essential for effective TTA}. 

(v) Semantic-NeRF which fuses  segmentation masks across views in a geometrically consistent manner outperforms single view segmentation performance of \maskformerbsln{} by 3\%.\looseness=-1 

(vi) \model{} which adapts model parameters of the segmentor at test time greatly outperforms Semantic-NeRF in OOD scenes. 

(vii) \maskformerreconbsln{} performs worse with TTA, which suggests that the decoder's design is very important for aligning the reconstruction and segmentation tasks.

For qualitative comparisions with Mask2former and additional qualitative results, please refer to Figure \ref{fig:mask2former_compare} and Figure \ref{fig:fast_slow_suppl} in the Appendix. Further in Section \ref{sec:results_rgbmulti_sup} of the Appendix, we show  results on a different distribution shift, where we include object instances from novel object categories that are not present in the train set.  Additionally we also include results on multi-view CLEVR dataset.

\subsubsection{Synthesizing and Decomposing Unseen Viewpoints}
\label{sec:nvs}
We evaluate \model{}'s  ability to render and decompose novel (unseen) RGB image views. We consider the same dataset and train-test split as  above, where in the MSN-Hard dataset we have 5-7 object instances in the training set and 16-30 object instances in the test set.
Our model takes two views as input, and uses two views for TTA, and the remaining five (unseen) views are used to evaluate rendering quality.  We evaluate the pixel-accurate reconstruction quality PSNR and segmentation ARI accuracy for the remaining five novel unseen viewpoints, which are not seen during TTA training. Views are randomly sampled in the different sets.  We fit SemanticNeRF in the same 4 (input and target) views, and evaluate it in the same remaining five views as our model. 

We see in Table \ref{tab:novelview}  that \model's rendering quality on novel (unseen) viewpoints improves with test-time adaptation. This means  that test-time adaptation does not only improve segmentation on the test-time adapted  viewpoints, but also improves the view synthesis quality and segmentation accuracy on novel (unseen) viewpoints. Further, we find that Semantic-NeRF does not generalize as well to novel (unseen) viewpoints. NeRFs are well known to perform poorly with a small number of views \cite{yu2021pixelnerf,Johari_2022_CVPR}. This is because the model does not have any inductive biases of scene structure. Rather, an MLP renderer is optimized for each scene separately.

\subsection{Decomposing RGB images in single-view scenes}
\label{sec:singleview_rgb}
We  test \model{} in it's ability to segment single-view RGB images on CLEVR \cite{johnson2017clevr} and ClevrTex \cite{karazija2021clevrtex} which are standard object-centric learning datasets. We compare \model{} against state-of-the-art supervised and unsupervised scene decomposition methods such as \maskformerbsln{} \cite{mask2former} and Slot Attention \cite{locatello2020object}.

\paragraph{Dataset}
We consider CLEVR \cite{johnson2017clevr} and it's out-of-distribution variant ClevrTex \cite{karazija2021clevrtex}. For supervised training, we use the CLEVR dataset open-sourced by \citet{multiobjectdatasets19}. The train set consists of standard CLEVR scenes sampled from 3 object shapes, 8 object colors and 2 object materials. For the test set we use the ClevrTex dataset of \citet{karazija2021clevrtex} which is sampled from 8 object shapes and 85 object materials. Specifically we use their publicly available ClevrTex-PlainBG dataset, thus resulting in a significant distribution shift in terms of object properties. Both the datasets contain 3-10 object instances per scene.

\paragraph{Baselines}
We compare against the following baselines:

(i) Mask2Former \cite{mask2former} a state-of-the-art 2D image segmentor.

(ii) \model{}-w/o supervision, has the same model architecture as our method except is not trained using supervised segmentation loss.  This method is similar to Slot Attention \cite{locatello2020object}.

\paragraph{Results}
We show our results in Table \ref{tab:singleimage}.  
Our findings are similar to Section \ref{sec:multiview}, for instance \model{} significantly outperforms Mask2former on out-of-distribution scenes after doing test-time adaptation. Similarly \model{} gets similar results to Mask2former on the in-distribution set after adaptation. Further training Slot Attention with supervision is important to get high ARI accuracy.

\begin{table}[t]
\centering
\scalebox{0.73}{
\begin{tabular}{@{\extracolsep{0pt}}llcccc@{}}
\toprule
\multirow{2}{*}{\textbf{Method}}
&\multicolumn{2}{c}{\textbf{PSNR}} 
&\multicolumn{2}{c}{\textbf{ARI}} \\
\cmidrule{2-3} \cmidrule{4-5}
 & Direct Infer.  & with TTA.  & Direct Infer.  & with TTA.  \\
\midrule
Semantic-NeRF  & \quad N/A & 18.9 & N/A &  0.51 \\
\model{} (Ours)   & \quad 19.7 & \textbf{22.6} & 0.57 &  \textbf{0.68} \\
\bottomrule
\end{tabular}
}
\caption{ \textbf{RGB rendering and segmentation accuracy} (higher is better) in  out-of-distribution test set of 16-30 object instances. }
\label{tab:novelview}

\end{table}

\begin{table}[t]
\centering
\scalebox{0.7}{
\begin{tabular}{@{\extracolsep{-3pt}}llcccc@{}}
\toprule
\multirow{2}{*}{\textbf{Method}}
&\multicolumn{2}{c}{\textbf{in-dist: CLEVR}} 
&\multicolumn{2}{c}{\textbf{out-of-dist: ClevrTex}} \\
\cmidrule{2-3} \cmidrule{4-5}
 & Direct Infer.  & with TTA.  & Direct Infer.  & with TTA.  \\
\midrule
\model{}-w/o supervision    & \quad 0.21 & 0.22 & 0.15 &  0.37 \\ 
\midrule
\maskformerbsln    & \quad \textbf{0.97} & N/A & 0.64 &  N/A \\ 
\model{} (Ours)   & \quad 0.95 & \textbf{0.97} & 0.35 &  \textbf{0.68} \\
\bottomrule
\end{tabular}
}
\caption{ \textbf{Instance Segmentation ARI accuracy}  (higher is better) for single-view RGB images.   Out-of-distribution scenes are sampled from ClevrTex having different object shapes and materials compared to the in-distribution train set of CLEVR.}
\label{tab:singleimage}

\end{table}

\subsection{Decomposing  3D point clouds}
\label{sec:exppointcloud}
We test \model{} in its ability to segment 3D object point clouds into parts.
We consider two types of distribution shifts: (i) Part-to-object distribution shift, where  our model and baselines  are supervised from a dataset of generic 3D part primitives and are tested on segmenting 3D object point clouds. (ii) Cross-object-category distribution shift, where our model and baselines are supervised from 3D object part segmentations and tested on segmenting instances of \textit{novel (unseen) categories} into parts. 

\subsubsection{Part-to-object distribution shift} 
\label{sec:generic}

\paragraph{Dataset}
We consider the dataset split of Shape2Prog \citep{shap2prog}. Our supervised training set consists of scenes that contain  2-3 primitive parts, resized and translated in different 3D locations of a blank 3D canvas. The part primitives (akin to  generalized cylinders of \citet{Marr}) consist of differently sized cubes, cuboids, and discs. Our test set consists of unseen object categories specifically chairs and tables from PartNet, each composed of 6 to 16 parts. 

\textbf{Baselines \quad} We consider the following semi-supervised 3D baselines which show results for the aforementioned train-test split in their respective papers: 

(i) PQ-Nets of  \citet{Wu_2020_CVPR}, which  assumes access to a set of primitive 3D parts for pre-training. Specifically they first learn a primitive part decoder, then they learn a sequential encoder that encodes the 3D point cloud into a 1D latent vector and sequentially decodes parts using the pretrained part decoder. We use the publicly available code to train the model. 

(ii) Shape2Prog of \citet{shap2prog}, which is a shape program synthesis method that is trained supervised to predict shape programs from object 3D point clouds. The program represents the part category, location, and the symmetry relations among the parts (if any).

 We further consider  the following ablative versions  of \model{}: 
 
(i) \model{} w/o supervision, a  variant of our model that does not use any segmentation supervision.

(ii) \model{} w/o SlotAttention, which instead of Slot Attention it maps 3D point features to slots via iterative layers of cross (query to point) and self (query-to-query) attention layers on learnable query vectors similar to  DETR \citep{carion2020end}. Please note that slots and queries represent the same thing, but we use the terminology of DETR \citep{carion2020end} in this case.

(iii) \model{} w/o SlotDecoder, \textit{does not use a information bottleneck during mask decoding}. Rather it decodes the mask by computing a dot product between slot vectors and  the feature grid, similar to \maskd{}. For the reconstruction head it uses the same decoder as \model{}.

\begin{table}[h]
\centering
\scalebox{0.72}{
\begin{tabular}{@{\extracolsep{-1pt}}llcccc@{}}
\toprule
\multirow{2}{*}{\textbf{Method}}
&\multicolumn{2}{c}{\textbf{in-dist:} 
\includegraphics[width=0.45cm]{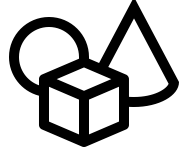}  

} 
&\multicolumn{2}{c}{\textbf{out-dist:
\includegraphics[width=0.45cm]{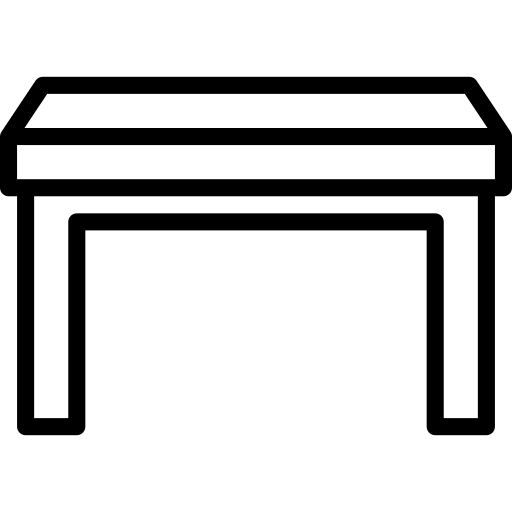}  
\includegraphics[width=0.45cm]{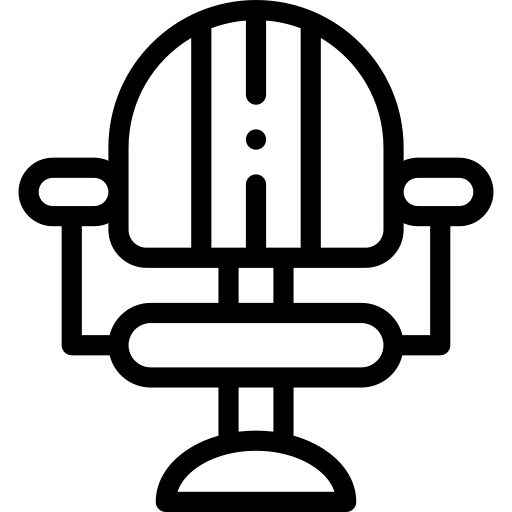}  
}} \\
\cmidrule{2-3} \cmidrule{4-5}
 & Direct Infer.  & with TTA.  & Direct Infer.  & with TTA.  \\
\midrule

Shape2Prog   &  \quad0.65  & 0.71  & 0.26 & 0.38 \\ 
PQ-Nets   & \quad 0.63 & 0.67 & 0.19 &  0.28 \\ 
\midrule
\model{} w/o SlotAttention  & \quad0.71 & 0.74  & 0.41 & 0.52 \\ 
\model{} w/o Supervision  & \quad 0.42 & 0.38  & 0.38 & 0.27 \\ 
\model{} w/o SlotDecoder  & \quad\textbf{0.79} & \textbf{0.77}  & 0.42 & 0.33 \\ 
\midrule
\model{} (Ours)  & \quad0.69 & 0.75 & \textbf{0.44} &  \textbf{0.58} \\
\bottomrule

\end{tabular}

}
\caption{\small \textbf{Instance Segmentation ARI accuracy} 
(higher is better) in instances from generic primitives dataset (in-distribution) and Chair and Table categories (out-of-distribution) when trained using the supervision from generic primitive compositions (part-to-object distribution shift).}
\label{tab:unsupervised}
\end{table}

\begin{figure*}[t!]
    \begin{center}
    \includegraphics[width=1.0\textwidth]{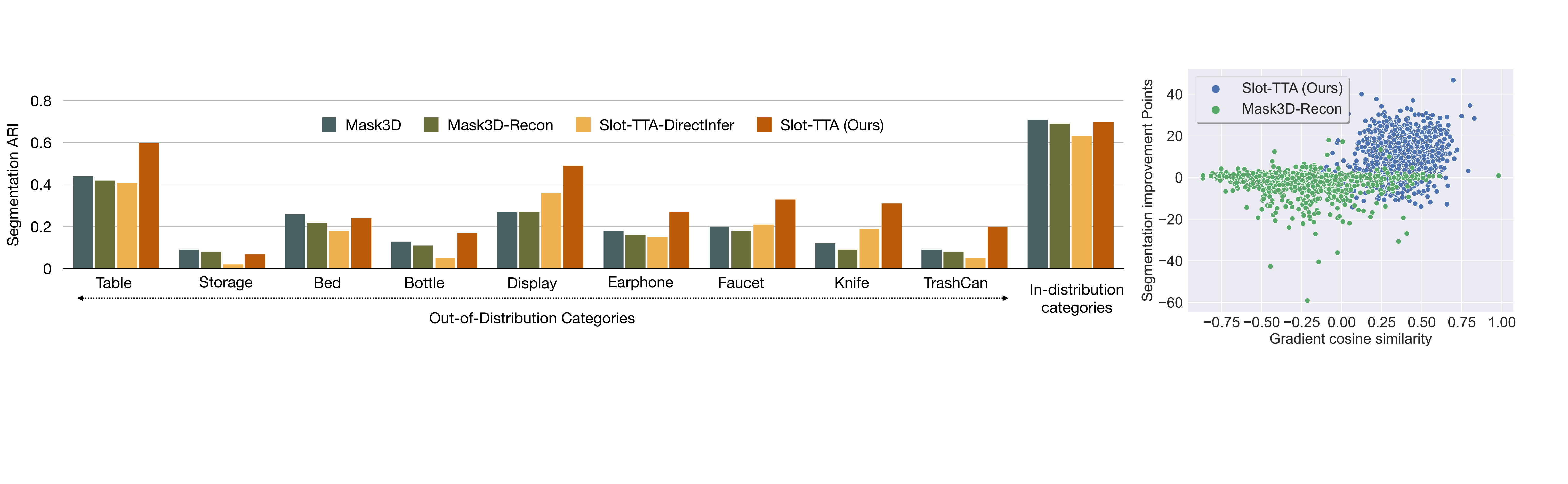}
    \end{center}
    \vspace{-0.5cm}
    \caption{\textbf{Left:} Instance segmentation ARI accuracy on out-of-distribution categories for \model{} and baselines.
    \textbf{Right:} Segmentation-reconstruction gradient similarity versus segmentation improvement. We plot the cosine similarity of gradients from reconstruction and segmentation losses for various examples in the test-set paired with their corresponding difference in segmentation accuracy before and after TTA (Improvement Points).
    }
  \label{fig:semisupervised_results}
\end{figure*}

\paragraph{Results}
We show quantitative  results of our model and  baselines in Table \ref{tab:unsupervised}.  
Our conclusions are as follows:

(i) \model{} significantly outperform PQ-Nets \citep{Wu_2020_CVPR} and Shape2Prog \citep{shap2prog}. 

(ii) \model{} outperforms \model{} w/o SlotAttention, which indicates that competition among slots is important for TTA.

(iii) Test-time adaptation through reconstruction feedback increases 3D part segmentation accuracy on both our model \textit{and} our baselines. We think this is due to the common slot bottleneck present in the baselines. However, the difference is that the baselines infer slot vectors one at a time using sequential RNN operations whereas our model infers them jointly using SlotAttention.

(iv) \model{} w/o supervision does not improve during TTA, much like in the multi view RGB image case of Section \ref{sec:multiview}.

(v) The direct inference version of \model{} w/o SlotDecoder significantly outperforms \model{} in distribution,  however it fails to improve with TTA in out-of-distribution setting. This suggests that  information bottleneck in the decoder is a plus for test-time adaptation, however it is a minus for fitting to the training distribution.

Please, refer to Section \ref{sec:generic_sup} of the Appendix for further ablations and qualitative comparison against baselines. Finally, please refer to Appendix Figure \ref{fig:generic_prims} for visualization of the 3D primitive dataset.

\subsubsection{Cross-category distribution shift}
\label{sec:semantic}

\paragraph{Dataset}
We divide object categories in the PartNet benchmark \citep{mo2019partnet} into train and test sets such that there is no overlap of categories between the two sets. Specifically, the model has access to the ground-truth  point cloud segmentation of eight categories and is tested on the remaining 9 PartNet categories. We use annotation for the finest segmentation level available (level 3).

\paragraph{Baselines}
We compare our model against  the following baselines:  

(i) Mask-3D \citep{schult2022mask3d}, a state-of-the-art 3D instance segmentation model, which is a 3D adaptation of the  2D state-of-the-art segmentor \maskformerbsln{}.

(ii)  Mask-3D-Recon, where we add a reconstruction  decoding head to the Mask-3D baseline, for doing test-time adaptation.

Please refer to Section \ref{sec:baselines} of the Appendix for additional details on the baselines.

\paragraph{Results}
We show quantitative results in Figure \ref{fig:semisupervised_results} on the \textit{left} side and qualitative results in Figure \ref{fig:semisupervised} and in the Appendix.  
Our conclusions are as follows:

(i) \model{} outperforms Mask3D in  OOD categories and has comparable performance within the training distribution. 

(ii) Mask3D-Recon does not improve during test-time adaptation. This suggests that the decoder's architecture in \model{} are important for aligning the segmentation and reconstruction objectives. 
To illustrate this point further, we visualize in Figure  \ref{fig:semisupervised_results} on the \textit{right} side,

the cosine similarity between gradient vectors of segmentation  and  reconstruction loss during TTA of multiple individual scene examples  for our model and for the Mask3D-Recon baseline. As can be seen, higher gradient similarity correlates with larger improvements in segmentation performance from reconstruction gradient descent during TTA. We find that \model{} has  higher gradient similarity than Mask3D-Recon, which explains their performance difference during TTA. 

(iii) Direct inference in \model{} has a lower performance than \model{}, Mask3D and Mask3D-Recon baselines. 

\section{Discussion - Limitations} \label{sec:lim} 
As can be inferred from the results in Sections \ref{sec:multiview}, \ref{sec:generic} and \ref{sec:semantic}, the direct inference versions of baseline models such as Mask2former or Mask3D, significantly outperforms the direct inference version of \model. On the other hand, \model{} after TTA significantly outperforms the above baselines on out-of-distribution examples. This stark difference in performance between  direct inference and after TTA setting can be attributed to the information bottleneck of the segmentation decoder, which we ablate in Section \ref{sec:generic}. 
We find that the presence of an information bottleneck within the decoder adversely impacts the direct inference performance. This effect is found to escalate exponentially when dealing with complex datasets like MS-COCO ~\cite{lin2014microsoft}, even when trained supervised using human-annotated segmentations. This significantly diminished direct inference capability compromises the model's potential for doing test-time adaptation. Exploration of architectures that can both fit on large scale training data, such as COCO, and be test time adapted is a direct avenue of our future work. Our present work sheds lights to limitations and opportunities of slot-centric models when combined with entity segmentation supervision.\looseness=-1

\vspace{-0.1in}
\section{Conclusion} \label{sec:conclusion}
We presented \model{}, a novel semi-supervised scene decomposition model equipped with a slot-centric image or point-cloud rendering component for test time adaptation. We showed 
\model{} greatly improves instance segmentation on out-of-distribution scenes using test-time adaptation on reconstruction or novel view synthesis objectives. We compared with numerous baseline methods, ranging from state-of-the-art feedforward segmentors, to NERF-based TTA for multiview semantic fusion, to state-of-the-art TTA methods, to unsupervised or weakly supervised 2D and 3D generative models. We showed \model{} compares favorably against all of them for scene decomposition of OOD scenes, while still being competitive within distribution.%

\textbf{Acknowledgements} We thank Mike Mozer, Klaus Greff and Ansh Khurana for the helpful discussions and Aravindh Mahendran for the paper
feedback. Part of the Work was done during Mihir Prabhudesai's internship at Google. This material is based upon work supported by DARPA Young Investigator Award, a NSF CAREER award, 
DARPA Machine Common Sense, ONR N00014-22-1-2096 and ONR MURI N00014-22-1-2773.

\bibliography{7_refs}
\newpage
\section*{Appendix}
The structure of this appendix is as follows:
In Section ~\ref{sec:dataprep} we cover the details on the datasets. In Section~\ref{sec:impl} we specify further implementation details. In Section~\ref{sec:results_sup} we provide additional qualitative and quantitative results for the experiments in Section \ref{sec:experiments} of our main paper.

\section{Datasets}\label{sec:dataprep}
\subsection{Multi-view RGB}\label{sec:multiview_suppl}
\begin{figure}[ht!]
  \centering
    \includegraphics[width=1.0\textwidth]{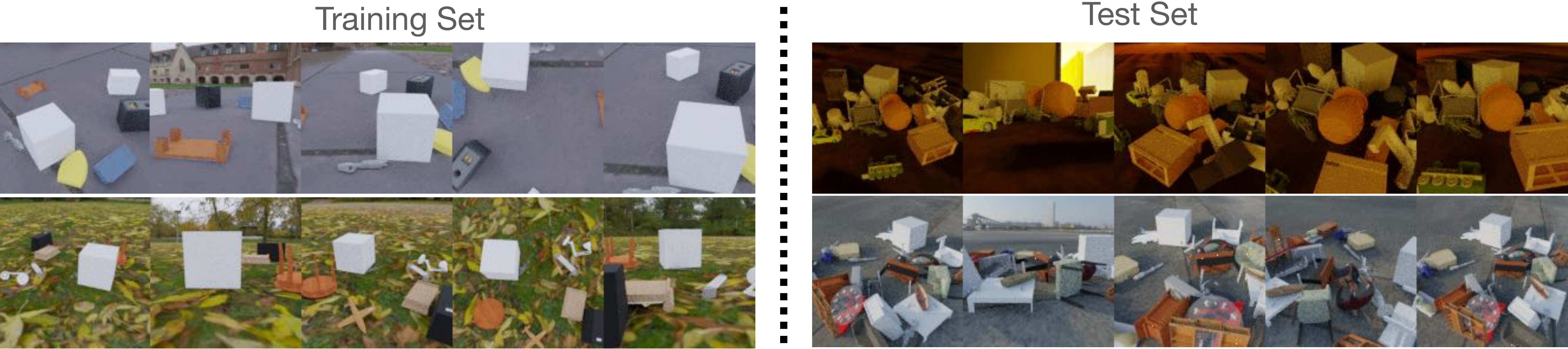}
  \caption{We visualize samples from the train-test split used by us in experiment Section \ref{sec:multiview}. Different rows correspond to different scenes and different columns correspond to different viewpoints.}
    \label{fig:msn_train_test}
\end{figure}

  We use the MultiShapeNet-Hard dataset of Scene Representation Transformer, a complex photo-realistic dataset for Novel View Synthesis~\citep{sajjadi2022scene}. Our train split consists of 5-7 ShapeNet objects placed at random locations and orientations in the scene. The backgrounds are sampled from 382 realistic HDR environment maps.  Our test set consists of 16-30 novel object instances placed at novel arrangements. We sample objects from a pool of 51K ShapeNet objects across all categories, we divide the pool into train and test such that the test set consists of objects not seen during training. The train split has 200K scenes, and the test set consists of 4000 scenes, each with 9 views.
  We had to regenerate the dataset for this specific train-test split.

\subsection{Single-view RGB}\label{sec:single_view_suppl}

We use the CLEVR dataset of \citet{johnson2017clevr}, which includes RGB images and segmentation masks rendered using Blender. For the training set we use the official dataset opensourced by \citet{multiobjectdatasets19}. For the test-set we use the official dataset of ClevrTex by \citet{karazija2021clevrtex}. Specifically we use their publicly available ClevrTex-PlainBG dataset. Due to computational cost of TTA, we only use the first 1000 scenes in the dataset for testing.

\subsection{Point Cloud}\label{sec:pointcloud_suppl}
\subsubsection{Generic primitive part dataset.}
\label{fig:generic_suppl}
\begin{figure}[ht!]
  \centering
    \includegraphics[width=0.7\textwidth]{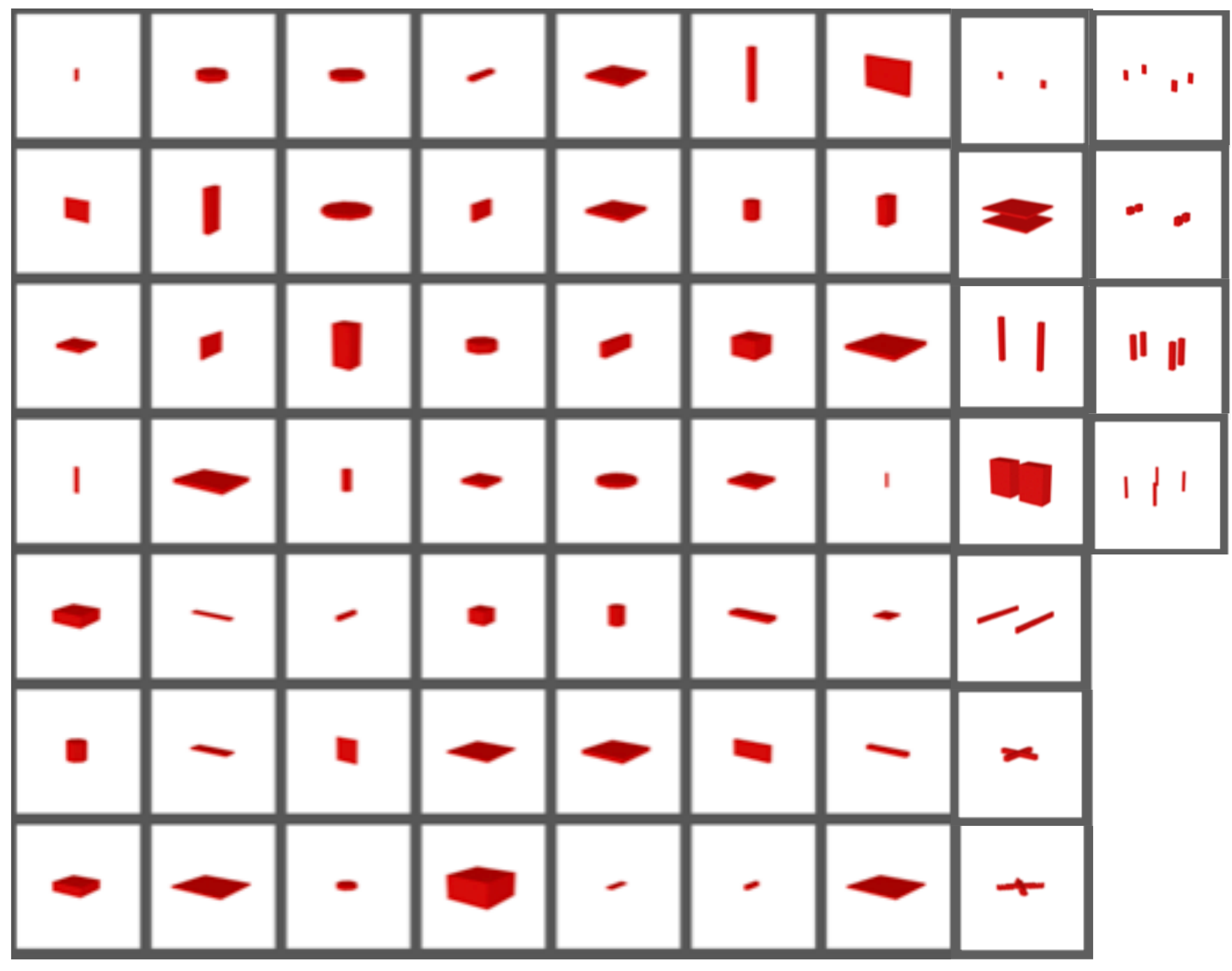}
  \caption{We visualize all the generic primitive templates of \cite{shap2prog}, as you can see, they mainly consist of Cubes, Cuboids, and Discs.}
    \label{fig:generic_prims}
\end{figure}
We use the primitive dataset of \cite{shap2prog} as supervision in Experiment Section \ref{sec:generic}. The dataset consists of 200K primitive instances sampled from the primitive templates that are visualized in Figure \ref{fig:generic_prims}. Examples are sampled from the templates by changing their sizes and placing 2-3 primitives uniformly in random locations. The primitives are represented using a 32x32x32 binary voxel grid.

\subsubsection{PartNet dataset.} 
We use the official level-3 train-test split of PartNet \citep{mo2019partnet}. We randomly split the categories in PartNet into train and test. Our train categories consist of: Chair, Lamp, Clock, Refrigerator, Microwave, Dishwasher, Door and Vase. Our test categories consist of:  Table, Storage, Bed, Bottle, Display, Earphone, Faucet, Knife and TrashCan. We use this as the train-test split in Experiment Section \ref{sec:semantic}. We set the value of number of slots $K$ as 16 for this dataset. We provide all 10K points as input to \model{} and baselines. For evaluation, we calculate ARI segmentation accuracy on occupied points after voxelizing 10K points into  a 32x32x32 binary voxel grid.

\section{Implementation details}\label{sec:impl}
\subsection{Posed multi-view 2D RGB images }
\paragraph{\normalfont \textbf{Training details and computational complexity.}}
We use a batch size of 256 in this setting. We set our learning rate as $10^{-4}$. We use an Adam optimizer with $\beta_1 = 0.9$, $\beta_2 = 0.999$. For training, our model takes about 4 days to converge using 64 TPUv2 chips. Test-time adaptation for each example takes about 10 seconds on a single TPUv2 chip. Similarly, a forward pass through our model takes about 0.1 seconds. During training, instead of decoding all the pixels,  we decode only a sample of them. Specifically, we randomly pick 1024-pixel locations for each example in the batch during each iteration of training.  During test-time adaptation, instead of uniformly sampling pixel locations, we use an error-weighted sampling strategy which we describe below. 

\paragraph{\normalfont \textbf{Inputs.}} 
During training and test-time adaptation, our model takes in as input multi-view RGB images along with their ground-truth egomotion.
For each scene, we randomly select four input and five target views.  We use a resolution of 128x128 for our input and target images.

\paragraph{\normalfont \textbf{Encoder.}}
Here we follow the original implementation of OSRT \citep{sajjadi2022scene}. The model encodes each input image $\I_i$, its camera extrinsic and intrinsics into a set representation via a shared CNN and transformer backbone. Specifically, the CNN outputs a feature grid for each image conditioned on the camera extrinsic and intrinsics, which are then flattened into a set of flat patch embeddings. The patch embeddings are then processed by a transformer that outputs a set of latent embeddings. The latent embeddings have a dimensionality of 1535. The CNN consists of 3 blocks of convolutions, with a ReLU activation after each convolution. The transformer contains 5 blocks of Multi-Head Self-attention. 

\paragraph{\normalfont \textbf{Slot Attention.}}
The latent embeddings from the encoder are then mapped to a Slot Attention module. We use the original implementation by \cite{locatello2020object}, however instead of initializing the slots from a multi-variate gaussian we have them as learnable embedding vectors. We keep our slot vectors dimensionality as 1536. We set the number of slots as 32 in this setting.

\paragraph{\normalfont \textbf{Decoder.}}
We use the broadcast decoder of \cite{sajjadi2022object} for decoding the slots to their RGB image conditioned on the target viewpoints. Our slot decoder consists of a 4-layer MLP with a hidden dimensionality of 1536 and ReLU activation.  Our target viewpoints are parameterized using 6D light-field parametrization of camera position and normalized ray direction.

\paragraph{Error-conditioned pixel sampling}
To accelerate test-time adaptation, we sparsely sample a subset of pixels from the target images, where we prioritize the pixels with a high reconstruction error. To this end, we calculate the reconstruction error over all pixels and apply a Softmax with a temperature $\tau = 0.01$ along the pixel dimension.

\subsection{3D point clouds}
\paragraph{\normalfont \textbf{Training details and computational complexity.}}
 We use a batch size of 16 for point cloud input. We set our learning rate as $40^{-4}$. We use the Adam optimizer with $\beta_1 = 0.9$, $\beta_2 = 0.999$.
Our model takes 24 hours (approximately 200k iterations) to converge. Our test-time adaptation per example takes about 1 min (500 iterations). A forward pass through the proposed model takes about 0.15 secs. We use a single V100 GPU for training and inference.

\paragraph{\normalfont \textbf{Inputs.}} 
We subsample the input point clouds of 10K points to a standard size of 2048 points, before passing it to the encoder.

\paragraph{\normalfont \textbf{Encoder.}}
We adopt the point transformer \citep{zhao2021point} architecture as our encoder. Point transformer encoder is essentially layers of self-attention blocks. Specifically, a self-attention block includes sampling of query points and updating them using their $N$ most neighboring points as key/value vectors. In the architecture, we apply 5 layers of self-attention which look as follows: 2048-16-64,  2048-16-64,  512-16-64,  512-16-64,   128-16-64, 128-16-64. We use the notation of $S$-$N$-$C$, where $S$ is the number of subsampled query points from the point cloud, $N$ is the number of neighboring points and $C$ is the feature dimension. We thus get an output feature map of size $128 \times 64$.

\paragraph{\normalfont \textbf{Decoder.}} 
We obtain point occupancies by querying the slot feature vector $slot_k$ at discrete locations $(x, y, z)$ specifically $o_{x,y,z} = \Dec(slot_k, (x,y,z))$. 
The architecture of $\Dec$ is similar to that of \cite{coconets}. Given $slot_k$, which is one of the slot feature vector. We encode the coordinate $(x, y, z)$ into a 64-D feature vector using a linear layer. We denote this vector as $z$.
The inputs $slot_k$ and $z$ are then processed as follows: 
$out_k = RB_i( z + FC_1(slot_k))\cdots)$
We set $i=3$. $FC_i$ is a linear layer that outputs a 64 dimensional vector. $RN_i$ is a 2 layer ResNet MLP block \citep{he2016deep}. The architecture of ResNet block is: ReLU, 64-64, ReLU, 64-64. Here, $i-o$ represents a linear layer, where $i$  and $o$ are the input and output dimension. Finally $out_k$ is then passed through a ReLU activation function followed by a linear layer to generate a single value for occupancy.

\subsection{Baselines}\label{sec:baselines}

\paragraph{\normalfont \textbf{Mask2former} \citep{mask2former}}
Mask2former is a recent state-of-the-art 2D RGB segmentation network, that scales transformer-based DETR ~\citep{carion2020end} for the task of segmentation. They improve DETR's transformer decoder by adding masked and multi-scale attention, which helps them achieve SOTA results on  panoptic, instance and semantic segmentation on the COCO dataset. We use their publicly available code to train on the MultiShapeNet-Hard dataset. We use a batch size of 256 and train their network on 8 V100s GPUS for four days until convergence. We set the number of slots in their network as 32, similar to our model.

\paragraph{\normalfont \textbf{Mask2former-BYOL}}
Following the implementation of MT3 \cite{bartler2022mt3}, we add a byol head on top of the slot vectors Mask2former. Specifically, we compress the slot vectors into a single vector also commonly known as <CLS TOKEN> in ViT\cite{dosovitskiy2020image}. We then follow the implementation of MT3 where add a BYOL head on top of this vector. We use all the augmentations originally used by Mask2former for computing the non-contrastive loss.

\paragraph{\normalfont \textbf{Mask2former-Recon}}
Similar to \model{}, we add an RGB decoder on top of the slot vectors of Mask2former. Specifically, we use the same implicit broadcast decoder and alpha composting of \model{} to predict the scene RGB. Note that we only predict the RGB and not the segmentation mask from this decoder.

\paragraph{\normalfont \textbf{Mask3D.} \citep{schult2022mask3d}}
Mask3D is a 3D re-implementation of Mask2former~\citep{mask2former}(a state-of-the-art object 2D segmentation method) for the task of instance segmentation. Mask3D doesn't officially show any results on PartNet dataset, so we adapt their code to fit the resolution of PartNet dataset while keeping their core architecture the same. 

\paragraph{\normalfont \textbf{Mask3D-Recon}}
We follow the same design choice as Mask2former-Recon, however, we add the reconstruction decoder to Mask3D instead of Mask2former.

\paragraph{\normalfont \textbf{Shape2Prog} \citep{shap2prog}} Shape2Prog is a shape program synthesis method that is trained supervised to predict shape programs from object 3D point clouds. 
Shape2Prog introduced two synthetically generated datasets that helped the model parse 3D pointclouds from ShapeNet \cite{shapenet2015} into shape programs without any supervision: i) Generic primitive set (Figure \ref{fig:generic_prims})  we discussed earlier in which they use to pre-train their part decoders. Shape2Prog assumes access to a Synthetic whole shape dataset of chairs and tables generated programmatically alongside its respective ground-truth programs. Their model requires supervised pre-training on the dataset of synthetic whole shapes paired with programs. Inorder to maintain the OOD shift, we don't assume access to synthetic whole shapes dataset, however instead we train their encoder to predict programs/segment multiple instances of primitive parts. We use their open-sourced code for comparision with our model. We change the value of number of blocks similar to the number of slots in our model.

\paragraph{\normalfont \textbf{PQ-Nets.} \citep{Wu_2020_CVPR}}
PQ-Nets is a sequential encoder-decoder architecture, that takes 3D point cloud as input and sequentially encodes it into multiple 1D latents which are then decoded to part point clouds. It achieves this decomposition by pre-training their decoder to predict part point clouds. We use their open-sourced architecture and code for comparision with our model. We train their model using our datasets from scratch. We change the value of number of slots in their model based on the maximum number of parts in the dataset.

\section{Additional Experiments}
\label{sec:results_sup}
\subsection{Segmenting  RGB images in multi-view scenes}
\label{sec:results_rgbmulti_sup}

\begin{table}[h]
\centering
\scalebox{0.6}{
\begin{tabular}{@{\extracolsep{4pt}}llcccc@{}}
\toprule
\multirow{2}{*}{\textbf{Method}}
&\multicolumn{2}{c}{\textbf{in-dist (5-7 instances)}} 
&\multicolumn{2}{c}{\textbf{out-of-dist (16-30 instances)}} \\
\cmidrule{2-3} \cmidrule{4-5}
 & Direct Infer.  & with TTA.  & Direct Infer.  & with TTA.  \\
\midrule
\model{}-SlotMixer\_Decoder     & \quad \textbf{0.94} & 0.89 & 0.65 &  0.72 \\ 
\model{}-SRT\_Decoder    & \quad 0.92 & 0.88 & 0.60 &  0.63 \\ 
\midrule
\model-tta\_All\_param  & \quad N/A & 0.92  & N/A & 0.82 \\ 
\model-tta\_Norm\_param   & \quad N/A & 0.94 & N/A &  0.79 \\
\model-tta\_Slot\_param   & \quad N/A & 0.94 & N/A &  0.76 \\
\midrule  
\model{} w/o Weighted\_Sample   & \quad N/A & 0.93 & N/A &  0.81 \\
\midrule  
\model{} (Ours)   & \quad 0.92 & \textbf{0.95} & \textbf{0.70} &  \textbf{0.84} \\
\bottomrule
\end{tabular}
}
\caption{ \textbf{ Instance Segmentation ARI accuracy} (higher is better) in the in-distribution test set of 5-7 object instances and out-of-distribution 16-30 object instances. 
}
\label{tab:osrt2}

\end{table}

\begin{table}[h]
\centering
\scalebox{0.65}{
\begin{tabular}{@{\extracolsep{4pt}}llcccc@{}}
\toprule
\multirow{2}{*}{\textbf{Method}}
&\multicolumn{2}{c}{\textbf{in-dist (ShapeNet categories)}} 
&\multicolumn{2}{c}{\textbf{out-of-dist (GSO categories)}} \\
\cmidrule{2-3} \cmidrule{4-5}
 & Direct Infer.  & with TTA.  & Direct Infer.  & with TTA.  \\
\midrule
\maskformerbsln & \quad 0.93 & N/A & \textbf{0.93} & N/A \\
\maskformerbyolbsln & \quad 0.93 & 0.95 & 0.92 & 0.93 \\
\maskformerreconbsln & \quad 0.93 & 0.92 & 0.92 & 0.91 \\

\midrule  

\model{} (Ours)   & \quad 0.92 & \textbf{0.95} & 0.92 &  \textbf{0.95} \\
\bottomrule
\end{tabular}
}
\caption{ \textbf{Instance Segmentation ARI accuracy} (higher is better) in the in-distribution test set of ShapeNet object categories\cite{shapenet2015}  and out-of-distribution test set of GSO object categories \cite{downs2022google}. 
}
\label{tab:gso}

\end{table}

In Table \ref{tab:gso}, we tested our model on a different distribution shift. In the test set instead of increasing the number of instances in the scene in Table \ref{tab:osrt}, we introduced instances from new object categories. Specifically the MSN \cite{sajjadi2022scene} train-set consists of ShapeNet object categories\cite{shapenet2015} (Tables, Chairs etc), whereas the new test-set consists of Google Scanned Object \cite{downs2022google} (GSO) categories (Shoes, Stuffed toys etc). 

In Table \ref{tab:clevr}, we tested our model on CLEVR dataset of \citep{johnson2017clevr} using the same train-test setup as Section \ref{sec:multiview}, where in the train set we use 4-7 objects and in the test set we use 7-10 objects. As can be seen \model{} achieves close to perfect results out-of-dist shift after TTA.

\begin{table}[h]
\centering
\scalebox{0.70}{
\begin{tabular}{@{\extracolsep{4pt}}llcccc@{}}
\toprule
\multirow{2}{*}{\textbf{Method}}
&\multicolumn{2}{c}{\textbf{in-dist (4-7 instances)}} 
&\multicolumn{2}{c}{\textbf{out-of-dist (7-10 instances)}} \\
\cmidrule{2-3} \cmidrule{4-5}
 & Direct Infer.  & with TTA.  & Direct Infer.  & with TTA.  \\
\midrule
\model{} (Ours)   & \quad 0.96 & \textbf{0.97} & 0.92 &  \textbf{0.97} \\
\bottomrule
\end{tabular}
}
\caption{ \textbf{ Instance Segmentation ARI accuracy} (higher is better) in the in-distribution test set of 4-7 object instances and out-of-distribution 7-10 object instances of CLEVR dataset. 
}
\label{tab:clevr}
\end{table}

\begin{figure}[H]
  \centering
    \includegraphics[width=1.0\textwidth]{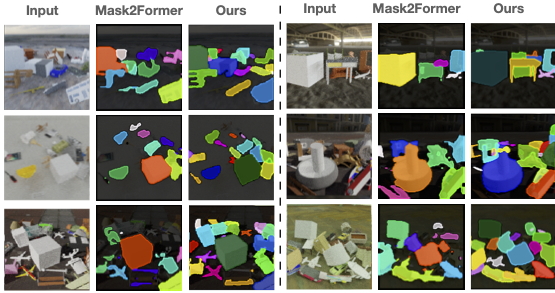}
  \caption{We compare Mask2former with \model{}-with TTA on out-of-dist test set of 16-30 object instances, following the setup of Section \ref{sec:multiview}.}
    \label{fig:mask2former_compare}
\end{figure}

\begin{figure}[H]
  \centering
    \includegraphics[width=1.0\textwidth]{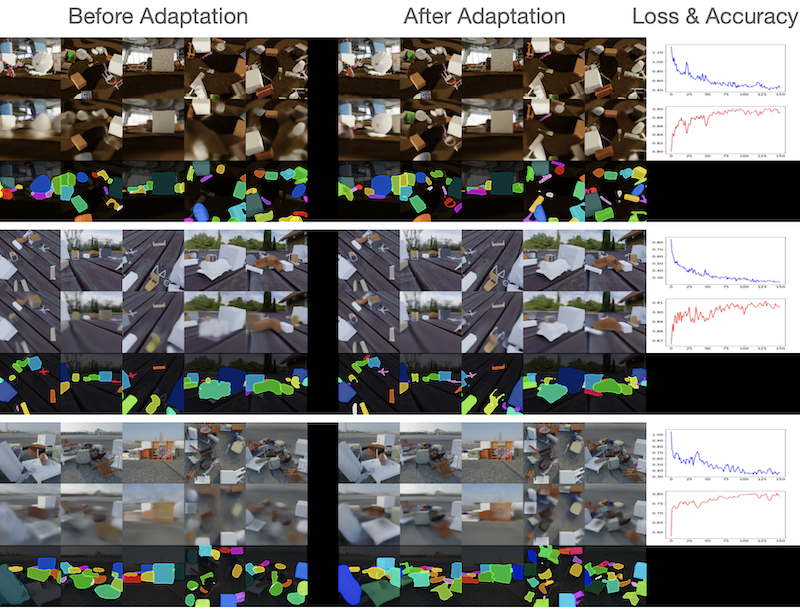}
  \caption{On the left, we visualize \model{}-DirectInfer. In the middle, we visualize \model{}-with TTA. In the first row we visualize the ground truth target RGB views. In the second and third row we visualize \model{} predicted target RGB views and their segmentation masks. On the right-most column we visualize the RGB loss and segmentation accuracy during  adaptation.}
    \label{fig:fast_slow_suppl}
\end{figure}

We conduct various ablations of \model{} considering the same setting as \ref{sec:multiview} in Table \ref{tab:osrt}. 
In Figure \ref{fig:fast_slow_suppl}, we show additional qualitative results comparing \model-DirectInfer and \model-with TTA. In Figure \ref{fig:mask2former_compare}, we qualitatively compare Mask2former with \model{}.

(i) We ablate different decoder choices in the topmost section where instead of using the broadcast decoder we use the Scene representation transformer (SRT) decoder~\citep{sajjadi2022scene} which we refer to as \textbf{\model{}-SRT\_Decoder } or the SlotMixer decoder~\citep{sajjadi2022object}, referred to as \textbf{\model{}-SlotMixer\_Decoder}.

(ii)  We ablate  what parameters to adapt at test time. As it's unclear since TENT ~\citep{wang2020tent} optimizes BatchNorm or LayerNorm parameters, but TTT ~\citep{sun2020test} optimizes the shared parameters between the SSL and the task-specific branch, which in our case will be all the parameters in the network. 
In Table \ref{tab:osrt2}, \textbf{\model-tta\_All\_param} is when we adapt all the network parameters, \textbf{\model-tta\_Norm\_param} 
adapts only the Layer or BatchNorm parameters and \textbf{\model-tta\_Slot\_param} adapts only the learnable slot embeddings. We find that optimizing only the encoder parameters works the best for our setting.

(iii) Further, we ablate error-conditioned pixel sampling where \textbf{\model{} w/o Weighted\_Sample} refers to our model that uses uniform sampling instead of the error weighted sampling. 
\newpage

\subsection{Decomposing 3D point clouds}
\label{sec:generic_sup}

\begin{figure}[h!]
  \centering
    \includegraphics[width=1.0\textwidth]{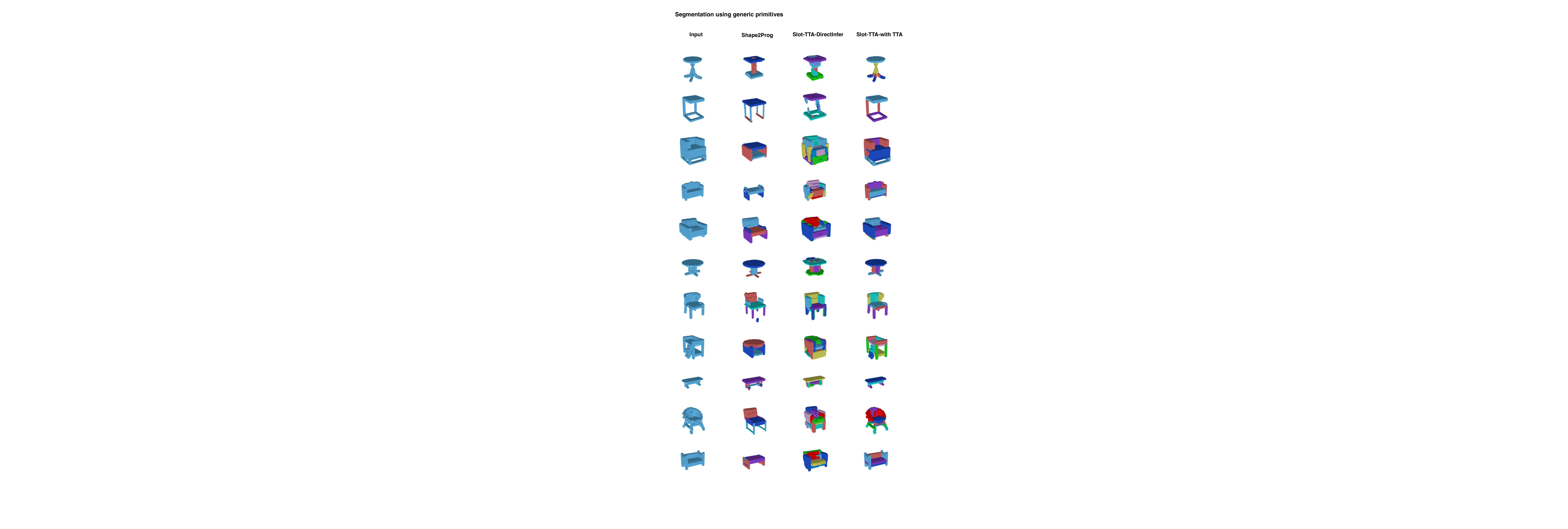}
  \caption{Additional segmentation results on out-of-distribution categories when supervised from generic primitives. Same setting as Section \ref{sec:generic}}
    \label{fig:generic_fig}
\end{figure}

We show additional qualitative results for Section \ref{sec:generic} in Figure \ref{fig:generic_fig}.

\bibliographystyle{icml2023}



\end{document}